\documentclass[conference]{IEEEtran}
\IEEEoverridecommandlockouts

\usepackage{cite}
\usepackage{amsmath,amssymb,amsfonts}
\usepackage{graphicx}
\usepackage{textcomp}
\usepackage{xcolor}
\usepackage{color}
\usepackage{url}
\usepackage{multirow}
\usepackage{booktabs}
\usepackage{algorithm,algpseudocode}
\DeclareMathOperator*{\argmin}{arg\,min}

\def\BibTeX{{\rm B\kern-.05em{\sc i\kern-.025em b}\kern-.08em
    T\kern-.1667em\lower.7ex\hbox{E}\kern-.125emX}}
\begin{document}

\title{TACL: Threshold-Adaptive Curriculum Learning Strategy for Enhancing Medical Text Understanding\\
\thanks{This work is supported by Provincial Natural Science Foundation of Jiangsu (NO. BK20250741, NO. BK20240703) , and the Startup Foundation for Introducing Talent of NUIST (NO. 1523142401057, 1523142401055).}
}

\author{
    Mucheng Ren, Yucheng Yan, He Chen, Danqing Hu, Jun Xu, and Xian Zeng$^{\dagger}$~\thanks{Corresponding author: xianzeng@nuist.edu.cn}\\
    \textit{Jiangsu Key Laboratory of Intelligent Medical Image Computing, School of Artificial Intelligence}\\
    Nanjing University of Information Science and Technology, Nanjing, China\\
    \{renm, 202412491503, chenh, danqinghu, jxu, xianzeng\}@nuist.edu.cn
}

\maketitle

\begin{abstract}
Electronic medical records (EMRs) are crucial for modern healthcare, containing rich information about patient care, diagnoses, and treatments. However, their unstructured nature, domain-specific language, and complexity pose significant challenges for automated understanding. Existing methods often treat all data equally, limiting their ability to handle rare or complex cases effectively. We present TACL (Threshold-Adaptive Curriculum Learning), a novel framework that dynamically adjusts the training process based on sample complexity. Inspired by progressive learning, TACL categorizes data into difficulty levels, focusing on simpler cases early in training and gradually addressing more complex ones. A domain-specific pre-trained language model is used for difficulty assessment, considering semantic, syntactic, and contextual features. Additionally, TACL employs an adaptive training strategy to enhance task-specific performance and ensure generalization across diverse datasets. Experimental results on multilingual datasets, including MIMIC-III, MIMIC-IV, and Chinese clinical records, demonstrate TACL's effectiveness in tasks such as ICD coding, readmission prediction, and TCM syndrome differentiation. TACL improves performance on rare and complex cases, providing a scalable and robust solution for medical text understanding.
\end{abstract}


\begin{figure}[t]
    \includegraphics[width=\linewidth]{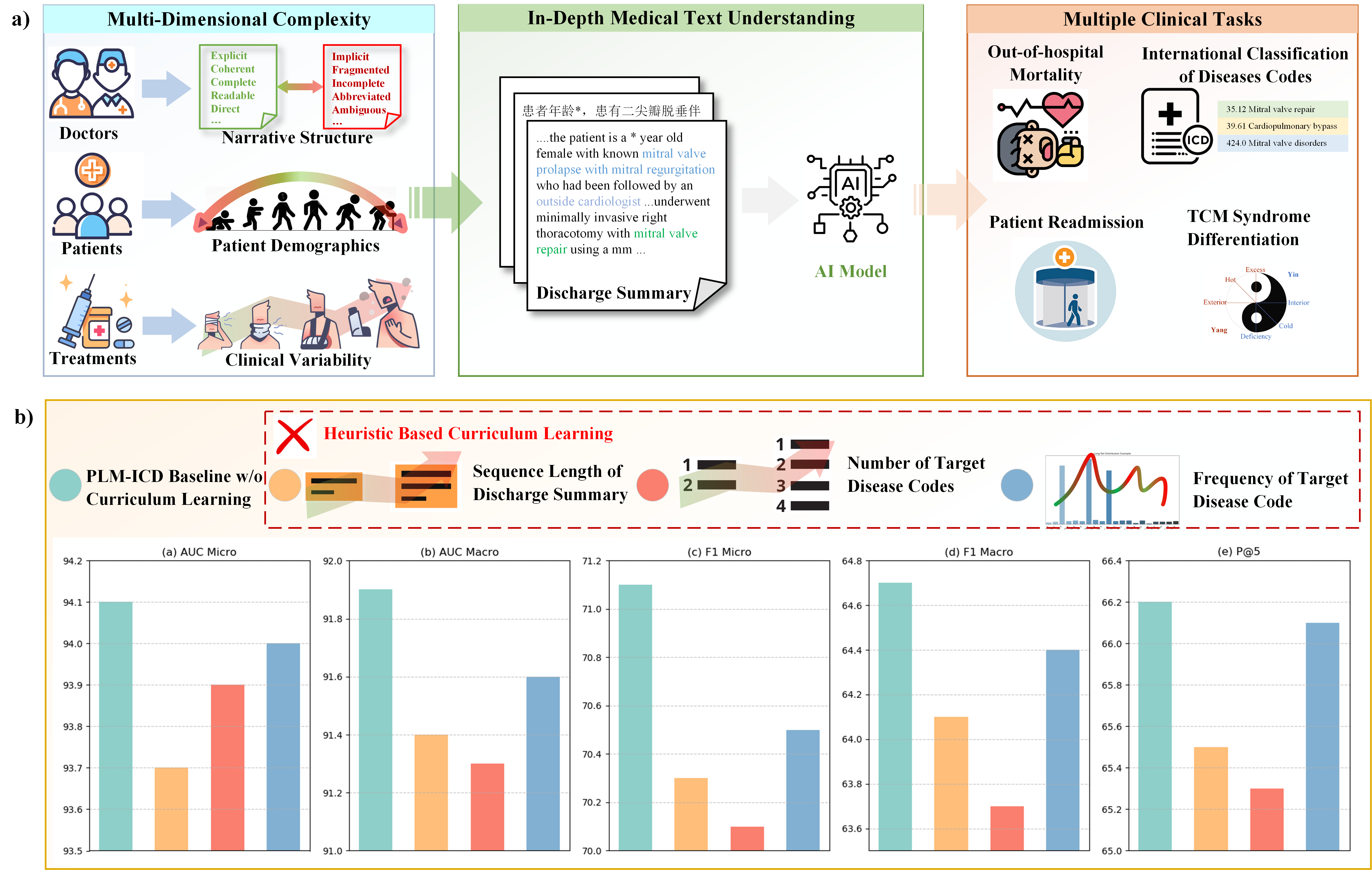}
    \caption{a) Illustration of Multi-Dimensional Complexity of Medical Text and Various Clinical Tasks relying on deep medical text understanding. b) A comparison of the performance of PLM-ICD on MIMIC-III-50-ICD9 without curriculum learning and with heuristic-based curriculum learning. }
    \label{fig:intro}
\end{figure}

\section{Introduction}
Medical texts, such as electronic medical records (EMRs), are essential for modern healthcare, capturing patient care, diagnoses, and treatments~\cite{giuffre2023harnessing}. However, their unstructured, domain-specific nature and variability—from straightforward cases to complex ones involving rare conditions—make automated understanding challenging.

As shown in Figure~\ref{fig:intro}(a), EMRs exhibit multi-dimensional complexity, including \textbf{narrative structure}, \textbf{clinical variability}, and \textbf{patient demographics}. Traditional NLP models follow a \textbf{“one-size-fits-all”} approach, often failing on rare and complex cases~\cite{edin2023automated,gomes2024accurate,lu2023towards,ren2022hicu,vu2021label,zhao2024automated}. This motivates a key question: \textbf{can models systematically learn from easier to harder cases, akin to human learning?}

Curriculum learning~\cite{bengio2009curriculum}, a training paradigm that progresses from simpler to more complex samples, has shown promise in healthcare applications like medical image segmentation~\cite{wang2023grenet,wang2023curriculum} and ICD coding~\cite{ren2022hicu}. However, key challenges remain: 1) \textbf{How to effectively assess the difficulty of medical texts?} and 2) \textbf{How to develop adaptive strategies to generalize across diverse tasks and datasets?} Simple metrics like sequence length or code frequency may fail to capture semantic nuances, as illustrated in Figure~\ref{fig:intro}(b), necessitating more robust methods.

To address these challenges, we propose TACL (\textbf{T}hreshold-\textbf{A}daptive \textbf{C}urriculum \textbf{L}earning), a framework designed to enhance medical text understanding. TACL introduces a novel difficulty assessment method leveraging a domain-specific pre-trained language model to quantify text complexity based on semantic, syntactic, and contextual features. Furthermore, it dynamically adjusts the curriculum to optimize learning across tasks and datasets.

TACL is validated on multilingual datasets, including English (MIMIC-III~\cite{johnson2016mimic}, MIMIC-IV~\cite{johnson2020mimic}) and Chinese clinical records, achieving state-of-the-art results in tasks like readmission prediction, out-hospital mortality, ICD coding, and TCM syndrome differentiation~\cite{ren2022tcm}. TACL excels in rare and complex cases, advancing medical text understanding.

\subsection*{Contributions}
\begin{itemize}
    \item \textbf{First application of curriculum learning in medical text understanding:} TACL introduces curriculum learning to handle the complexity and variability of clinical narratives, offering a novel perspective for this domain.
    \item \textbf{Difficulty assessment with domain-specific models:} A pre-trained language model quantifies medical text complexity, enabling systematic difficulty categorization for curriculum learning.
    \item \textbf{Adaptive training strategy:} TACL dynamically adjusts the learning process, enhancing generalization across diverse medical datasets and improving robustness in real-world applications.
\end{itemize}

\section{Related Work}
Advancements in NLP have addressed challenges in automated medical text understanding, particularly in \textbf{curriculum learning} and \textbf{medical text understanding}.

\subsection{Curriculum Learning in NLP}
Curriculum learning~\cite{bengio2009curriculum} structures training by starting with simpler examples and progressing to harder ones. Dynamic methods~\cite{platanios-etal-2019-competence,huang2020curricularface,kong2021adaptive} adapt difficulty during training but focus on general NLP tasks, overlooking domain-specific challenges like medical texts. TACL bridges this gap with a threshold-adaptive mechanism tailored to medical data, dynamically prioritizing samples to enhance task performance and generalization.

\subsection{Medical Text Understanding}
Early rule-based approaches and traditional machine learning models required extensive feature engineering. Deep learning, including CNNs, RNNs, and transformer-based models like BERT~\cite{devlin2019bert}, ClinicalBERT~\cite{huang2019clinicalbert}, and BioBERT~\cite{lee2020biobert}, improved medical NLP by leveraging pretraining on biomedical corpora. However, these models treat all samples equally, limiting their effectiveness on rare or complex cases. TACL addresses this by integrating curriculum learning to progressively train on simpler to harder cases, improving performance and generalization.

Datasets like MIMIC-III~\cite{johnson2016mimic} and MIMIC-IV~\cite{johnson2020mimic} highlight the variability in medical data, demonstrating the need for adaptive frameworks like TACL to handle heterogeneous and complex distributions.
\begin{figure*}[t]
    \centering
    \includegraphics[width=1\textwidth]{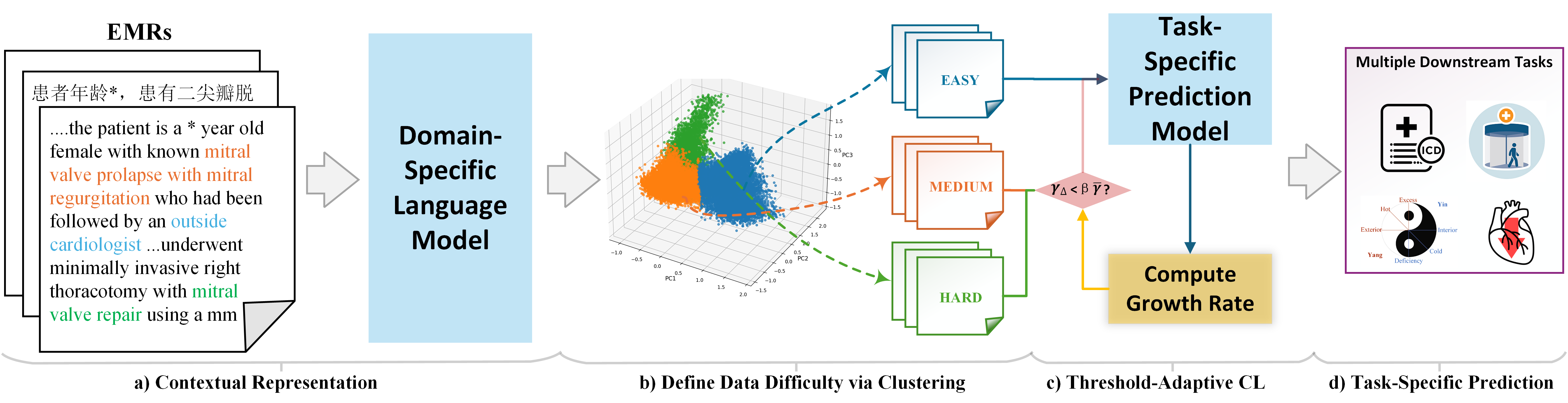}
    \caption{Overview of our proposed TACL framework.}
    \label{fig:overview}
\end{figure*}

\section{Method}

This section introduces our proposed framework, illustrated in Figure~\ref{fig:overview}, which consists of four key stages: (1) generating contextual representations using domain-specific pre-trained models, (2) defining data difficulty levels via clustering, (3) dynamically adjusting the curriculum with the Threshold-Adaptive Curriculum Learning (TACL) strategy, and (4) applying task-specific prediction heads for downstream tasks. These steps progressively guide the model through increasingly complex patterns within the data, ensuring robust learning across diverse medical NLP challenges.

\subsection{Contextual Representation}

To capture the semantic and terminological complexity of medical texts, we utilize domain-specific pre-trained models to generate high-dimensional contextual embeddings. Let the input dataset be denoted as:
\begin{equation}
\mathcal{D} = \{(x_i, y_i)\}_{i=1}^N
\end{equation}
where $x_i$ represents the input text, $y_i$ is the corresponding task-specific label (e.g., ICD code, readmission risk, or TCM syndrome), and $N$ is the total number of samples.
\begin{itemize}
    \item \textbf{English Tasks:} For tasks such as ICD coding out-of-hospital mortality, and readmission prediction, we employ \textbf{RoBERTa}~\cite{lewis-etal-2020-pretrained} pre-trained on medical corpora. 
    \item \textbf{Chinese Tasks:} For TCM syndrome differentiation, we utilize \textbf{ZY-BERT}~\cite{ren2022tcm} pre-trained on traditional Chinese medicine datasets. 
\end{itemize}

Thus, for each input text $x_i$, we extract the contextualized embedding of the \textbf{[CLS]} token, which serves as a summary representation of the entire input sequence. This embedding is obtained as follows:  
\begin{equation}
h_i = \text{Encoder}(x_i)_{[CLS]}, \quad h_i \in \mathbb{R}^d
\end{equation}
Here, $h_i$ represents the output embedding corresponding to the \textbf{[CLS]} token, which is widely used as the aggregate representation of the input sequence in transformer-based models. These embeddings form the foundation for subsequent clustering and curriculum learning steps.
\subsection{Data Clustering and Difficulty Definition}

To implement curriculum learning, we define difficulty levels by clustering the contextual representations $\{h_1, h_2, \dots, h_N\}$. This process ensures that the curriculum is aligned with the intrinsic structure of the data.

\paragraph{Clustering}
Given the embeddings $\{h_i\}_{i=1}^N$, we apply \textit{k-means clustering} to partition the samples into $k$ clusters, $\mathcal{C}_1, \mathcal{C}_2, \dots, \mathcal{C}_k$. The objective of k-means clustering is to minimize the within-cluster variance (WCSS):
\begin{equation}
\argmin_{\mu_1, \dots, \mu_k} \sum_{i=1}^k \sum_{h \in \mathcal{C}_i} \lVert h - \mu_i \rVert^2
\end{equation}
where $\mu_i \in \mathbb{R}^d$ is the centroid of cluster $\mathcal{C}_i$, defined as:
\begin{equation}
\mu_i = \frac{1}{|\mathcal{C}_i|} \sum_{h \in \mathcal{C}_i} h
\end{equation}

\paragraph{Difficulty Definition}
The difficulty of each cluster is determined by two factors:
\begin{itemize}
    \item \textbf{Cluster Density:} A measure of how tightly packed the embeddings are within a cluster. For cluster $\mathcal{C}_i$, the density is defined as:
    \begin{equation}
    \text{Density}(\mathcal{C}_i) = \frac{1}{|\mathcal{C}_i|} \sum_{h \in \mathcal{C}_i} \lVert h - \mu_i \rVert^2
    \end{equation}
    Lower density values indicate higher cluster density, meaning the samples are more homogeneous and less ambiguous.
    \item \textbf{Distance from Centroid:} The average distance of samples in a cluster from the centroid. For cluster $\mathcal{C}_i$, it is defined as:
    \begin{equation}
    \text{Distance}(\mathcal{C}_i) = \frac{1}{|\mathcal{C}_i|} \sum_{h \in \mathcal{C}_i} \lVert h - \mu_i \rVert
    \end{equation}
    Lower distance values indicate that the samples are closer to the centroid, representing a smaller and less diverse cluster.
\end{itemize}
Clusters are classified into three difficulty levels based on their density and distance values: \textit{Easy} clusters have low density values (high density) and low distance values, representing homogeneous and less diverse patterns; \textit{Medium} clusters have moderate density and distance values, reflecting patterns with some complexity and diversity but not overly challenging; and \textit{Hard} clusters have high density values (low density) and/or high distance values, representing highly diverse and ambiguous patterns.

\subsection{Threshold-Adaptive Curriculum Learning (TACL)}

Traditional curriculum learning approaches often rely on static or heuristic criteria for transitioning between difficulty levels. In contrast, our TACL strategy dynamically adjusts the curriculum based on the model's performance growth.

\paragraph{Growth Rate Metrics}
Let the macro-F1 score at epoch $t$ be denoted as $F_t$. We define two growth rate metrics to monitor the model's progress:
\begin{itemize}
    \item \textbf{Average Growth Rate:} Over a sliding window of the most recent $N$ epochs, the average growth rate is computed as:
    \begin{equation}
    \bar{\gamma} = \frac{1}{N-1} \sum_{t=2}^N \bigl(F_t - F_{t-1}\bigr)
    \end{equation}
    \item \textbf{Instantaneous Growth Rate:} The most recent growth rate is denoted as $\gamma_{\Delta}$ and is computed as:
    \begin{equation}
    \gamma_{\Delta} = F_N - F_{N-1}
    \end{equation}
\end{itemize}

\paragraph{Transition Condition}
A threshold coefficient $\beta \in (0, 1)$ is introduced to determine when the model has sufficiently learned from the current difficulty level. The model is considered \textit{saturated} at the current stage if:
\begin{equation}
\gamma_{\Delta} < \beta \cdot \bar{\gamma}
\end{equation}
When saturation is detected, the training progresses to the next difficulty stage.

\paragraph{Curriculum Update}
Let $\mathcal{D}_{\mathrm{easy}}, \mathcal{D}_{\mathrm{middle}}, \mathcal{D}_{\mathrm{hard}}$ represent the subsets of samples at each difficulty level. At each stage, training data is incrementally expanded:
\begin{align}
\mathcal{D}_{t+1} =
\begin{cases}
\mathcal{D}_{\mathrm{easy}}, & \text{if } \gamma_{\Delta} \geq \beta \cdot \bar{\gamma} \\[6pt]
\mathcal{D}_{\mathrm{easy}} \cup \mathcal{D}_{\mathrm{middle}}, & \substack{\text{if current stage = easy} \\ \text{and } \gamma_{\Delta} < \beta \cdot \bar{\gamma}} \\[6pt]
\mathcal{D}_{\mathrm{easy}} \cup \mathcal{D}_{\mathrm{middle}} \cup \mathcal{D}_{\mathrm{hard}}, & \substack{\text{if current stage = middle} \\ \text{and } \gamma_{\Delta} < \beta \cdot \bar{\gamma}}
\end{cases}
\end{align}
At the final stage (\textit{hard}), training continues until no improvement in macro-F1 is observed for $P$ consecutive epochs, after which training is terminated early.

\subsection{Downstream Task Prediction}
After training with TACL, the model is fine-tuned on multiple downstream tasks using task-specific prediction heads, optimized with cross-entropy loss for classification tasks and binary cross-entropy loss for multi-label tasks.

\begin{table}[t]
\centering
\caption{Dataset statistics for the tasks. The training dataset is divided into three difficulty levels: Easy, Medium, and Hard.}
\resizebox{\linewidth}{!}{%
\begin{tabular}{lcccccccc}
\hline
\multirow{2}{*}{Task} & \multirow{2}{*}{Language} & \multirow{2}{*}{Datasets} & \multicolumn{4}{c}{Train}                                        & \multirow{2}{*}{Val} & \multirow{2}{*}{Test} \\ \cline{4-7}
                      &                           &                           & Easy              & Medium            & Hard              & Total              &                      &                       \\ \hline
\multirow{2}{*}{Task 1: Readmission Prediction} 
                      & \multirow{2}{*}{English}  & Discharge Summary         & 12,256            & 8,472             & 5,517             & 26,245             & 3,027                & 3,063                 \\
                      &                           & Early Notes               & 22,858            & 13,196            & 11,739            & 47,793             & 5,774                & 5,441                 \\
Task 2: Out-of-hospital Mortality           
                      & English                   & MIMIC-IV                       & 14,437            & 12,823             & 6,227             & 33,487             & 4,186                & 4,186                 \\
                      
\multirow{3}{*}{Task 3: ICD  Coding}         
                      & \multirow{3}{*}{English}  & MIMIC-III-full            & 33,024            & 12,963            & 1,732             & 47,719             & 1,631                & 3,372                 \\
                      &                           & MIMIC-III-50              & 3,437             & 2,389             & 2,240             & 8,066              & 1,573                & 1,729                 \\
                       &                            & MIMIC-IV-ICD10            & 49,158            & 38,142            & 1,789             & 89,089             & 13,378               & 19,799                \\
Task 4: TCM Syndrome Differentiation            
                      & Chinese                   & TCM-SD                       & 33,386            & 5,114             & 4,680             & 43,180             & 5,486                & 5,486                 \\ \hline
\end{tabular}%
}
\label{tab:dataset_statistics}
\end{table}

\begin{figure*}
    \centering
    \includegraphics[width=\linewidth]{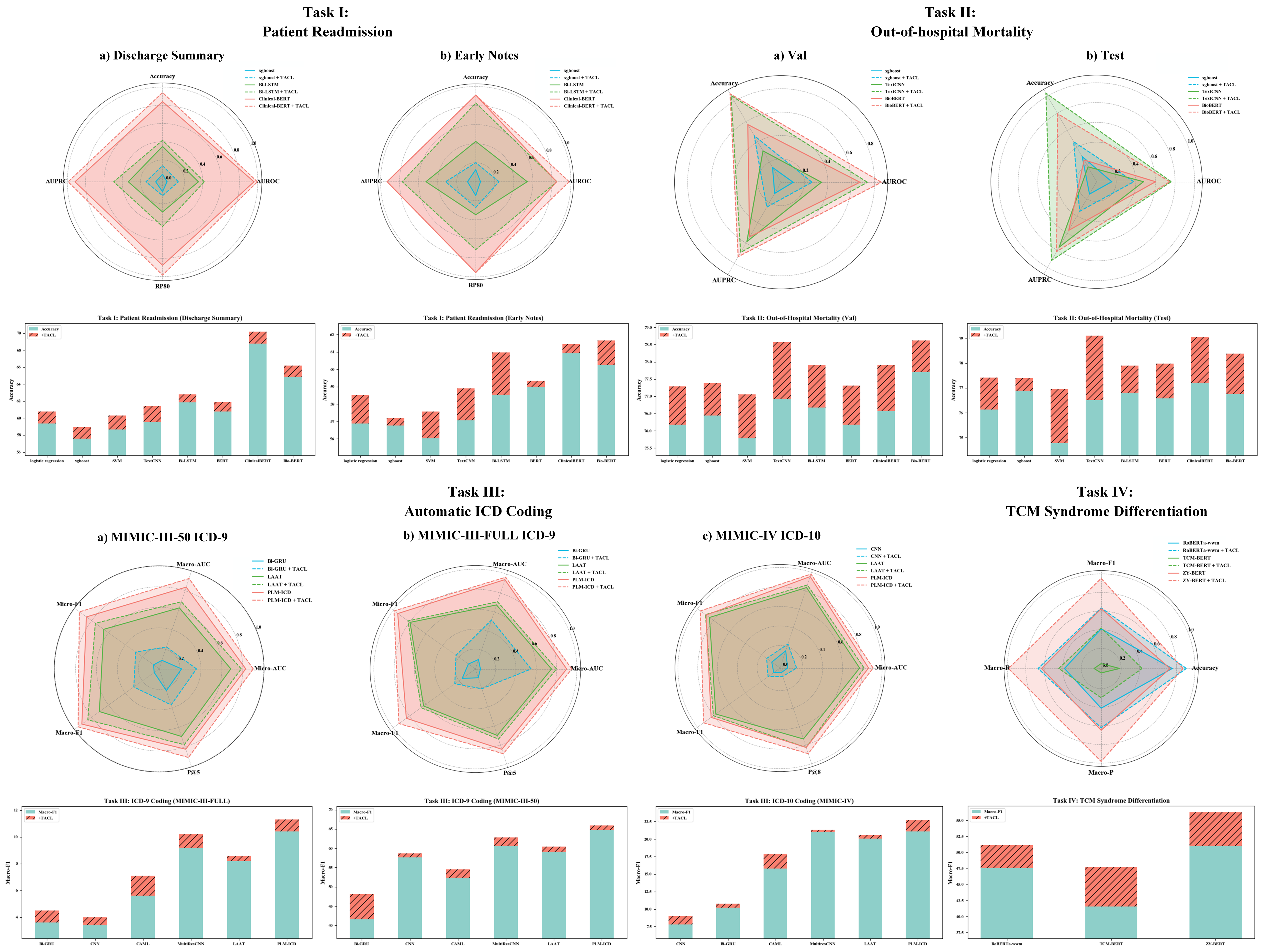}
    \caption{Performance comparison across four tasks: Patient Readmission (Task I), Out-of-hospital Mortality (Task II), ICD Coding (Task III), and TCM Syndrome Differentiation (Task IV).}
    \label{fig:main_results}
\end{figure*}
\section{Experiment}
\subsection{Datasets}
We evaluate TACL on four tasks using benchmark datasets partitioned into three difficulty levels (Easy, Medium, Hard). Table~\ref{tab:dataset_statistics} summarizes dataset statistics.

\textbf{Task 1: Patient Readmission:} Predicts if a patient is readmitted within a specific period using MIMIC-III~\cite{johnson2016mimic}. Two subsets are used:
\begin{itemize}
    \item \textbf{Discharge Summary:} Detailed discharge notes.
    \item \textbf{Early Notes:} Includes patient notes from 3 days earlier.
\end{itemize}

\textbf{Task 2: Out-of-hospital Mortality:} A binary classification task using MIMIC-IV~\cite{johnson2020mimic} discharge summaries to predict 30-day mortality post-discharge, excluding in-hospital deaths.

\textbf{Task 3: ICD Coding:} Multi-label classification on:
\begin{itemize}
    \item \textbf{MIMIC-III Full (ICD-9):} 52,723 documents, 8,929 codes (median: 14 codes/doc).  
    \item \textbf{MIMIC-III 50 (ICD-9):} 11,368 documents, top 50 codes (median: 5 codes/doc).  
    \item \textbf{MIMIC-IV (ICD-10):} Clinical notes annotated with granular ICD-10 codes.
\end{itemize}
Preprocessing includes lowercasing, removing non-alphanumeric characters, and truncation to 2,500–4,000 tokens.

\textbf{Task 4: TCM Syndrome Differentiation:} Classifies Traditional Chinese Medicine syndromes using TCM-SD~\cite{ren2022tcm}, a Chinese-language dataset with domain-specific terminology.

\subsection{Implementation}
Experiments are conducted on an NVIDIA L40 48GB GPU with fixed random seeds. TACL thresholds range from 0.5 to 0.8, optimized on validation subsets. Metrics include F1-score, AUC, and precision at K (P@K), computed on test sets using the best model on validation. All code will be publicly available\footnote{\url{https://anonymous.4open.science/r/medical-codes-354C}}.

\subsection{Baseline}  
We compare the proposed method against a range of strong baselines across all tasks. For tasks involving clinical and biomedical text, such as readmission prediction, out-of-hospital mortality, and ICD coding, we evaluate models like Bi-LSTM~\cite{schuster1997bidirectional}, which utilizes pre-trained Word2Vec embeddings and bidirectional LSTM layers, and transformer-based models like BioBERT~\cite{alsentzer2019publicly} and ClinicalBERT~\cite{huang2019clinicalbert}, both pre-trained on large biomedical corpora. For ICD coding, additional specialized models like CAML~\cite{mullenbach-etal-2018-explainable} with label attention mechanisms and MultiResCNN~\cite{li2020icd} for multi-scale feature extraction are employed, alongside PLM-ICD~\cite{huang-etal-2022-plm}, which integrates a pre-trained BERT encoder with a label attention mechanism. For TCM syndrome differentiation, we evaluate Chinese-language models like RoBERTa-wwm~\cite{cui2021pre}, pre-trained with Whole Word Masking on large-scale Chinese text, and ZY-BERT~\cite{ren2022tcm}, a TCM-specific model trained on a large unlabelled corpus, achieving state-of-the-art performance on TCM tasks.

\section{Main Results and Discussion}
\subsection{Main Results}
Figure~\ref{fig:main_results} summarizes TACL's performance across all tasks, showing consistent improvements over traditional training methods in metrics such as Macro-F1, AUROC, and RP80. 

TACL significantly enhances simpler models like Bi-GRU, CNN, and Bi-LSTM, particularly for readmission prediction and ICD coding, improving Macro-F1 scores and recall on rare labels. For advanced models like PLM-ICD, MultiResCNN, ClinicalBERT, and ZY-BERT, TACL provides smaller but consistent gains, especially for rare labels and imbalanced datasets, such as ICD-10 coding and TCM syndrome differentiation.

These results highlight TACL's adaptability and scalability across models and tasks, demonstrating its potential to improve multilingual and domain-specific medical text understanding.
\begin{table}[t]
\centering
\footnotesize 
\caption{Performance on Patient Readmission Prediction.}
\resizebox{\linewidth}{!}{
\begin{tabular}{cccccc|cccc}
\hline
\multirow{2}{*}{\textbf{Model}}               & \multirow{2}{*}{\textbf{Stragegy}} & \multicolumn{4}{c|}{\textbf{Discharge Summary}}                        & \multicolumn{4}{c}{\textbf{Early Notes}}                             \\ \cline{3-10} 
                                              &                                    & \textbf{AUROC} & \textbf{Accuracy} & \textbf{AUPRC} & \textbf{RP80}  & \textbf{AUROC} & \textbf{Accuracy} & \textbf{AUPRC} & \textbf{RP80}  \\ \hline
\multirow{4}{*}{\textbf{Logistic Regression}} & \textbf{Original}                  & 62.41          & 59.37             & 63.23          & 4.73           & 59.56          & 56.87             & 60.47          & 3.25           \\
                                              & \textbf{+Length}                   & 62.48          & 59.54             & 63.17          & 2.99           & 59.62          & 56.80             & 60.34          & 3.77           \\
                                              & \textbf{+TACL$\circlearrowleft$}                   & 63.60          & 59.98             & 64.88          & 5.10           & 60.19          & 57.22             & 60.62          & 4.56           \\
                                              & \textbf{+TACL$\circlearrowright$}                     & \textbf{64.81} & \textbf{60.78}    & \textbf{66.52} & \textbf{7.20}   & \textbf{61.68} & \textbf{58.50}    & \textbf{61.17} & \textbf{5.89}  \\ \hline
\multirow{4}{*}{\textbf{XGboost}}             & \textbf{Original}                  & 60.14          & 57.59             & 63.48          & 2.12           & 58.40          & 56.76             & 60.78          & 2.04           \\
                                              & \textbf{+Length}                   & 60.33          & 57.70             & 63.51          & 3.18           & 59.00          & 56.93             & 61.18          & 1.79           \\
                                              & \textbf{+TACL$\circlearrowleft$}                   & 61.35          & 57.92             & 64.01          & 3.00           & 59.21          & 56.88             & 61.09          & 2.86           \\
                                              & \textbf{+TACL$\circlearrowright$}                     & \textbf{62.12} & \textbf{58.93}    & \textbf{64.97} & \textbf{3.47}  & \textbf{59.96} & \textbf{57.20}    & \textbf{62.85} & \textbf{3.79}  \\ \hline
\multirow{4}{*}{\textbf{SVM}}                 & \textbf{Original}                  & 63.26          & 58.67             & 63.21          & 6.59           & 58.49          & 56.03             & 61.17          & 3.16           \\
                                              & \textbf{+Length}                   & 63.07          & 58.59             & 63.18          & 5.55           & 59.01          & 56.57             & 61.39          & 4.98           \\
                                              & \textbf{+TACL$\circlearrowleft$}                   & 63.77          & 59.09             & 64.02          & 6.88           & 58.97          & 57.02             & 62.71          & 5.50           \\
                                              & \textbf{+TACL$\circlearrowright$}                     & \textbf{64.29} & \textbf{60.29}    & \textbf{65.94} & \textbf{7.14}  & \textbf{60.41} & \textbf{57.57}    & \textbf{63.24} & \textbf{5.49}  \\ \hline
\multirow{4}{*}{\textbf{TextCNN}}             & \textbf{Original}                  & 63.79          & 59.56             & 66.71          & 4.65           & 60.15          & 57.06             & 62.37          & 6.46           \\
                                              & \textbf{+Length}                   & 63.65          & 59.49             & 66.53          & 3.10           & 60.11          & 56.99             & 62.16          & 5.56           \\
                                              & \textbf{+TACL$\circlearrowleft$}                   & 64.30          & 59.96             & 67.22          & 6.79           & 61.01          & 57.50             & 63.33          & 6.74               \\
                                              & \textbf{+TACL$\circlearrowright$}                     & \textbf{64.93} & \textbf{61.43}    & \textbf{68.04} & \textbf{6.13}  & \textbf{62.46} & \textbf{58.9}     & \textbf{64.21} & \textbf{7.16}  \\ \hline
\multirow{4}{*}{\textbf{Bi-LSTM}}             & \textbf{Original}                  & 65.62          & 61.87             & 67.67          & 9.02           & 62.33          & 58.54             & 64.71          & 4.86           \\
                                              & \textbf{+Length}                   & 66.49          & 61.87             & 68.07          & 12.26          & 63.39          & 59.31             & 65.91          & 7.99           \\
                                              & \textbf{+TACL$\circlearrowleft$}                   & 65.87          & 62.59             & 68.52          & 10.90          & 64.37          & 60.58             & 65.84          & 7.16           \\
                                              & \textbf{+TACL$\circlearrowright$}                     & \textbf{66.66} & \textbf{62.81}    & \textbf{69.90} & \textbf{13.97} & \textbf{64.82} & \textbf{60.96}    & \textbf{66.95} & \textbf{9.99}  \\ \hline
\multirow{4}{*}{\textbf{BERT}}                & \textbf{Original}                  & 66.37          & 60.76             & 68.89          & 8.46           & 62.67          & 58.99             & 64.55          & 3.52           \\
                                              & \textbf{+Length}                   & 66.56          & 60.29             & 68.68          & 7.66           & 62.07          & 58.91             & 64.37          & 4.08           \\
                                              & \textbf{+TACL$\circlearrowleft$}                   & 66.72          & 61.03             & 69.11          & 8.94           & 62.95          & 59.02             & 65.05          & 6.46           \\
                                              & \textbf{+TACL$\circlearrowright$}                     & \textbf{67.65} & \textbf{61.90}    & \textbf{70.01} & \textbf{11.61} & \textbf{63.40} & \textbf{59.33}    & \textbf{66.58} & \textbf{10.43} \\ \hline
\multirow{4}{*}{\textbf{Bio-BERT}}            & \textbf{Original}                  & 71.68          & 64.84             & 73.49          & 20.09          & 63.86          & 60.26             & 65.97          & 7.16           \\
                                              & \textbf{+Length}                   & 72.01          & 65.18             & 73.67          & 20.58          & 63.85          & 60.19             & 65.79          & 6.33           \\
                                              & \textbf{+TACL$\circlearrowleft$}                   & 71.85          & 64.92             & 73.51          & 20.32          & 63.99          & 60.47             & 66.01          & 7.89           \\
                                              & \textbf{+TACL$\circlearrowright$}                     & \textbf{72.52} & \textbf{66.18}    & \textbf{74.28} & \textbf{23.81} & \textbf{64.78} & \textbf{61.65}    & \textbf{66.78} & \textbf{10.77} \\ \hline
\multirow{4}{*}{\textbf{ClinicalBERT}}        & \textbf{Original}                  & 75.34          & 68.76             & 75.83          & 27.17          & 64.80          & 60.93             & 66.64          & 8.82           \\
                                              & \textbf{+Length}                   & 73.73          & 68.46             & 74.81          & 23.34          & 64.04          & 59.77             & 66.24          & 10.67          \\
                                              & \textbf{+TACL$\circlearrowleft$}                   & 74.93          & 70.26             & 74.80          & 20.04          & 64.24          & 60.72             & 66.05          & 9.48           \\
                                              & \textbf{+TACL$\circlearrowright$}                     & \textbf{75.93} & \textbf{70.16}    & \textbf{76.87} & \textbf{30.52} & \textbf{65.85} & \textbf{61.46}    & \textbf{68.38} & \textbf{13.36} \\ \hline
\end{tabular}
}
\label{tab:readmission_results}
\end{table}

\begin{figure}[t]
    \centering
    \includegraphics[width=1\linewidth]{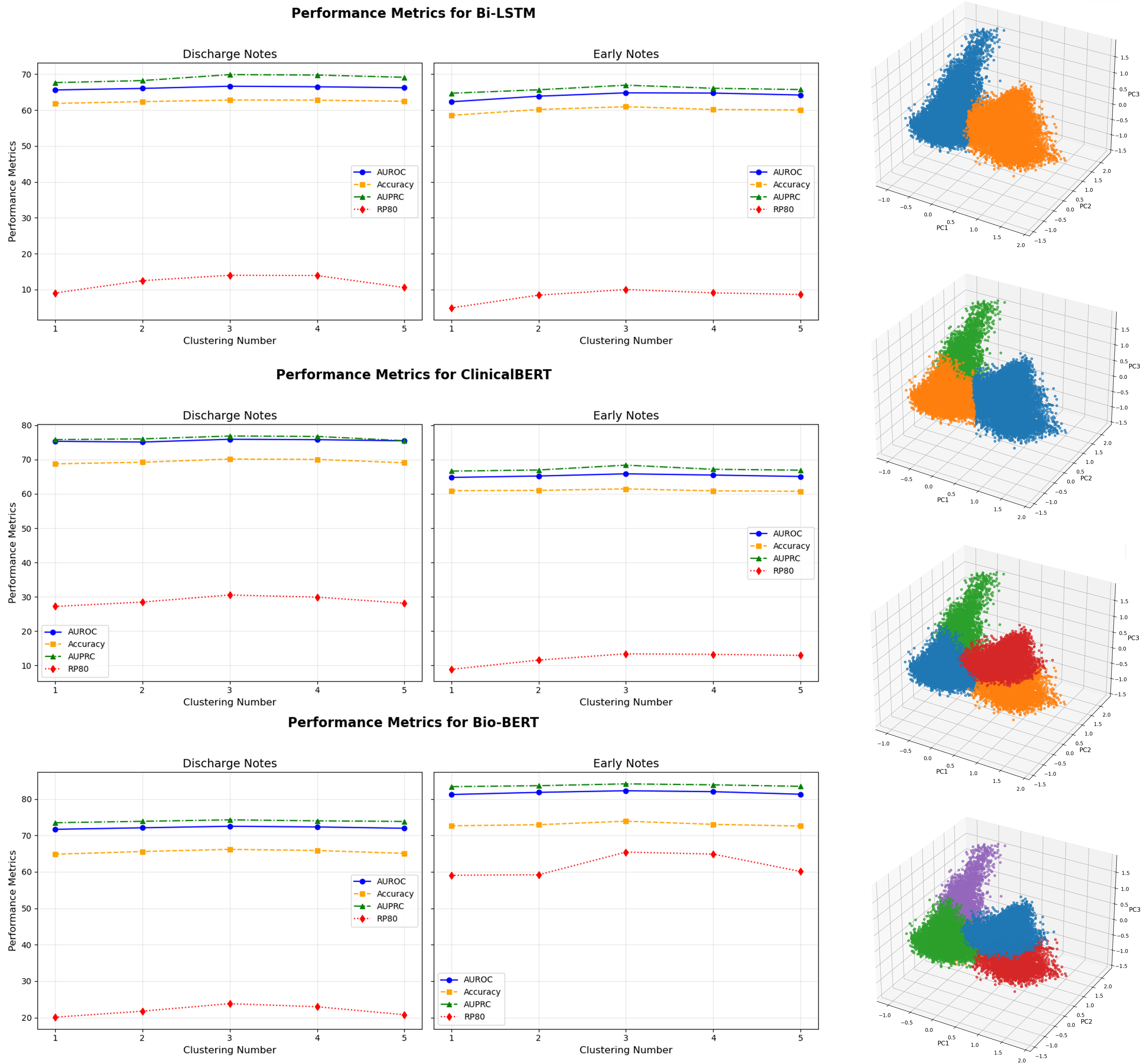}
    \caption{Ablation study and visualization about clustering number for readmission prediction task.}
    \label{fig:abla}
\end{figure}

\subsection{Analysis on Difficulty Definition}
Defining data difficulty is critical for curriculum learning. Traditional statistical features (e.g., sequence length, code frequency) degrade performance by disrupting sample ordering (Table~\ref{tab:readmission_results}). In contrast, contextual representations in TACL$\circlearrowright$ improve performance, with Bi-LSTM achieving +1.04 AUROC and +4.95 RP80 on discharge notes, highlighting the importance of task-aligned difficulty definitions.

\subsection{Analysis on Data Order}
Training from easy to hard ($\circlearrowright$) consistently outperforms hard-to-easy ($\circlearrowleft$) across all models and metrics (Table~\ref{tab:readmission_results}). Starting with simpler samples helps models build a foundation, enabling better handling of complex cases, whereas reversed order disrupts learning progression.

\subsection{Analysis on Clustering Number}
Clustering number impacts performance and data interpretability. Metrics peak at 3 clusters (Figure~\ref{fig:abla}), which balance data complexity and avoid over- or under-segmentation. Three clusters align well with inherent data structure, optimizing learning and performance.

\subsection{Clinical Applicability of TACL}
TACL enhances clinical applicability by improving prediction confidence and accuracy, particularly for rare labels and high-stakes tasks like mortality prediction. It adapts to diverse tasks and datasets, demonstrating consistent gains in AUROC, Macro-F1, and AUPRC across English and Chinese datasets. TACL’s scalability and ability to improve trust in AI predictions make it a valuable framework for advancing healthcare AI and improving patient outcomes.

\begin{figure}[t]
    \centering
    \includegraphics[width=1\linewidth]{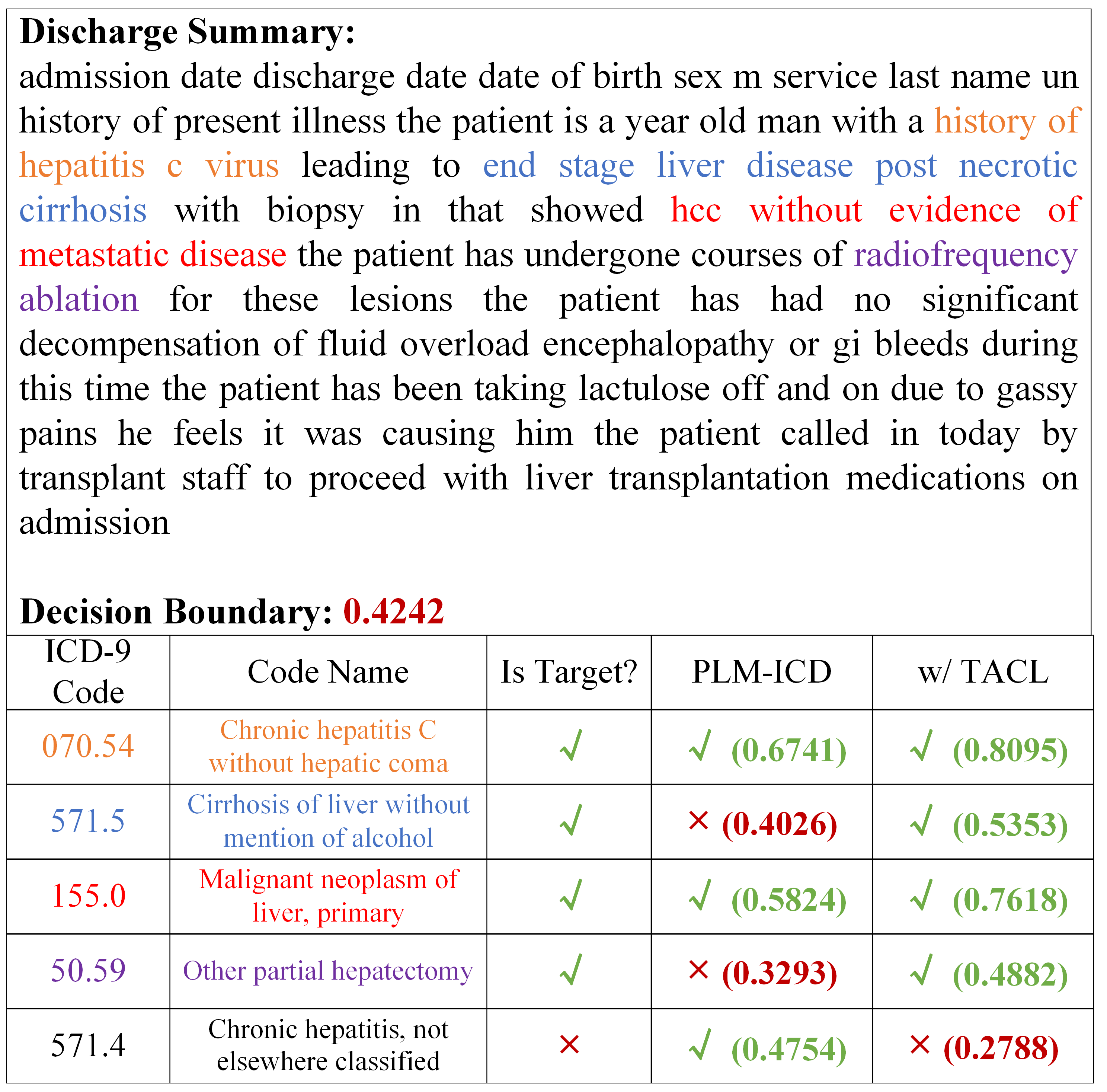}
    \caption{Case study of ICD-9 coding}
    \label{fig:case}
\end{figure}

\section{Conclusion}
This paper introduces TACL, a novel framework designed to address the nuanced challenges of medical text understanding by dynamically adapting the training process to data complexity. Using contextual representation-based difficulty definitions and a curriculum learning strategy that progressively transitions from simpler to more complex samples, TACL achieves significant improvements in key medical NLP tasks, including ICD coding, readmission prediction, and differentiation of TCM syndrome. The results demonstrate TACL’s ability to enhance model generalization, particularly in handling rare and complex cases, which are often the most clinically significant. Furthermore, the framework's adaptability to multilingual and domain-specific datasets, such as English clinical records and Chinese TCM texts, underscores its scalability and versatility. Beyond improving performance, TACL highlights the importance of aligning training strategies with task-specific nuances and data distributions, offering a transformative approach that bridges domain-specific challenges and scalable machine learning solutions.

\bibliographystyle{ieeetr}
\bibliography{mybibfile}

\end{document}